\documentclass{article}
\usepackage{spconf,amssymb,amsmath,epsfig}
\usepackage{graphicx}
\usepackage{multirow}
\usepackage{booktabs}
\usepackage{amssymb,amsmath,graphicx}
\usepackage{caption}
\usepackage{paralist}
\usepackage{subcaption}
\usepackage[export]{adjustbox}
\usepackage{moredefs} 
\usepackage{color}

\newcommand{\ZL}[1]{{\color{magenta} ZL: #1}}
\newcommand{\TS}[1]{{\color{blue} TS: #1}}

\usepackage{amssymb}

\usepackage{amsmath,amsfonts}
\usepackage{amsthm,amsopn}

\usepackage{bm} 

\newcommand{\vct}[1]{\boldsymbol{#1}} 

\newcommand{\field}[1]{\mathbb{#1}}
\newcommand{\R}{\field{R}} 
\newcommand{\T}{^{\textrm T}} 

\newcommand{\ProbOpr}[1]{\mathbb{#1}}

\newcommand{\expect}[2]{%
\ifthenelse{\equal{#2}{}}{\ProbOpr{E}_{#1}}
{\ifthenelse{\equal{#1}{}}{\ProbOpr{E}\left[#2\right]}{\ProbOpr{E}_{#1}\left[#2\right]}}} 
\newcommand{\var}[2]{%
\ifthenelse{\equal{#2}{}}{\ProbOpr{VAR}_{#1}}
{\ifthenelse{\equal{#1}{}}{\ProbOpr{VAR}\left[#2\right]}{\ProbOpr{VAR}_{#1}\left[#2\right]}}} 

\usepackage{enumitem}
\defcommand{\vec}[1]{\mathbf{#1}} 

\newcommand{\eat}[1]{}

\title{Learning Compact Recurrent Neural Networks}

\name{{Zhiyun Lu\thanks{*The author performed the work as a summer intern at Google, NYC.}$^{*1}$, Vikas Sindhwani$^{2}$, Tara N. Sainath$^{2}$}
\address{$^{1}$  University of Southern California, Los Angeles, CA, USA\\
$^{2}$ Google, Inc. New York, NY, USA \\
\fontsize{9}{9}\selectfont\ttfamily\upshape
zhiyunlu@usc.edu, 
\{sindhwani,tsainath\}@google.com}}

\begin{document}
\maketitle
\ninept
\begin{abstract}
Recurrent neural networks (RNNs), including long short-term memory (LSTM) RNNs, have produced state-of-the-art results on a variety of speech recognition tasks. However, these models are often too large in size for deployment on mobile devices with memory and latency constraints. In this work, we study mechanisms for learning compact RNNs and LSTMs via low-rank factorizations and parameter sharing schemes. Our goal is to investigate redundancies in recurrent architectures where compression can be admitted without losing performance. A hybrid strategy of using structured matrices in the bottom layers and shared low-rank factors on the top layers is found to be particularly effective, reducing the parameters of a standard LSTM by 75\%, at a small cost of 0.3\% increase in WER, on a 2,000-hr English Voice Search task.
\end{abstract}

\section{Introduction}

Recurrent neural network architectures have become very popular for automatic speech recognition (ASR) tasks in the past few years. Architectures such as recurrent neural networks (RNNs), long short-term memory networks (LSTMs) and convolutional, long-short term memory networks (CLDNNs) have produced state of the art results for many large vocabulary continuous speech recognition (LVCSR) tasks \cite{Saon14, Hasim14, Sainath15}.

In order to fully exploit thousands of hours of training data for LVCSR tasks, the best performing neural network architectures are typically very large in size. Consequently, they require long training time and consume a significant number of floating point operations per prediction once deployed.  These characteristics are further exacerbated in large, deep RNN architectures, which unroll the network for a sequential number of time frames (i.e., 10-20), and must compute the output from one time step before feeding into the next time step.  This situation is at odds with the need to enable high-performance on-device speech recognition on storage and power constrained mobile phones for which compact, small-sized models are strongly preferred.

Numerous approaches have recently been proposed in the model compression literature to build compact neural network models with fast training and prediction speed. A popular technique is low-rank matrix factorization \cite{Hasim14, tsainath:lowRank, xue2013restructuring}, which attempts to compress neural network layers by representing them as matrices with low-rank. This was shown to reduce parameters by 30-50\% for DNNs with no loss in accuracy for LVCSR tasks.  
Other techniques include inducing zeros in the parameter matrices via sparsity regularizers~\cite{CollinsKohli13}; storing weights in low fixed-precision formats~\cite{SuyogICML2015,LowPrecision2015};  using specific parameter sharing mechanisms~\cite{chen2015compressing, Sindhwani15}, or training smaller models on soft outputs of larger models~\cite{Distillation}. 

The dominant attention in this literature has so far been on reducing the size of fully connected and convolutional architectures (DNNs and CNNs).  Given the importance of recurrent architectures in the speech community, the goal of this work is to explore compact architectures for deep RNNs and LSTMs.  Several open questions immediately arise in this context: where precisely is the redundancy in recurrent architectures for speech recognition tasks? Which compact architectural variations retain the most performance? How should size constraints vary across layers, between recurrent and feedforward weights, and between different gates in a recurrent model? Our contributions in this paper are as follows:
\begin{compactitem}[$\bullet$]
\item We are the first to undertake a systematic study of various new compact architectures for RNNs and LSTMs. Specifically, we compare the effectiveness of low-rank models~\cite{Hasim14, tsainath:lowRank} and various parameter sharing schemes implemented using hashing~\cite{chen2015compressing} and structured matrices~\cite{Sindhwani15}. Note that sparsity promoting regularizers and the use of low-precision storage formats are complementary to our study and could yield even more compact models. 
\item Our investigation reveals the following: (i) for aggressive reduction of parameters in the bottom layers, Toeplitz-like structured matrices~\cite{Sindhwani15} outperform Hashing based schemes and low-rank factorizations, (ii) shared low-rank factors~\cite{Hasim14} are very effective for parameter reduction across the network, (iii) a particularly effective hybrid strategy for building compact LSTMs - using Toeplitz-like structured matrices~\cite{Sindhwani15} in bottom layers and projection layers involving shared low-rank factors in the upper layers -- can save 75\% parameters with 0.3\% increase in Word Error Rate (WER), compared to a full LSTM, (iv) LSTMs are relatively insensitive to whether compression is applied to recurrent or non-recurrent weights, and similarly so for input/output/forget gates; on the other hand, the cell state is critical to preserve for better performance.
\end{compactitem}

\section{Recurrent Neural Nets} \label{sec:models}
We start by setting some notations and by giving an overview of the basic RNN and LSTM architectures.

~\\
{\bf RNN}: An RNN  maps an input sequence $x = (x_1, \ldots, x_T)$ to output sequence $y = (y_1, \ldots, y_T)$. At each time step $t \in T$, the RNN is modeled with the following equations for the hidden unit $h^l_t$ in each layer $l \in \{1, \ldots, L\}$ 
and the output $y_t$, where $h^0_t = x_t$. 
\begin{align}
	h^l_t & = \sigma[W^lh^{l-1}_t + U^lh^l_{t-1} + b^l]  & l = 1, \ldots, L \notag \\
	y_t   & = \text{softmax}[W^{L+1}h^{L}_t + b^{L+1}]  & \label{full}
\end{align}
We will refer to Equation~\ref{full} as RNN full model. We call $U^l$ as recurrent weights, and $W^l$ as feedforward or non-recurrent weights. 

~\\
{\bf LSTM}: A LSTM is an alternative model that has been used to address vanishing/exploding gradient issues with RNN, and to model longer-term temporal dependencies~\cite{hochreiter1997long}. For each time step $t\in T$ and layer $l\in \{1, \ldots, L\}$, the LSTM sequence to sequence mapping, which includes the forget gate~\cite{gers2000learning} and peephole connection~\cite{gers2000recurrent}, is described as follows (as above $h^0_t = x_t$).
\begin{alignat}{3}	
	i^{l}_t  & = && \sigma[W_i^{l}h^{l-1}_t + U^{l}_ih^{l}_{t-1} + b^{l}_i + D(v^{l}_i)c^{l}_{t-1}]  				& \text{input gate} \label{input} \\
        f^{l}_{t}   & = && \sigma[W_f^{l}h^{l-1}_{t} + U^{l}_fh^{l}_{t-1} + b^{l}_f + D(v^{l}_f)c^{l}_{t-1}] 			& \text{forget gate}\label{forget} \\
        c^{l}_{t} & = && f^{l}_{t} \cdot c^{l}_{t-1} + i^{l}_t\tanh[W^{l}_ch^{l-1}_{t} + U^{l}_ch^{l}_{t-1} + b^{l}_c] 	& \text{cell state}\label{cell} \\
        o^{l}_t   & =&& \sigma[W^{l}_oh^{l-1}_{t} + U^{l}_oh^{l}_{t-1} + b^{l}_o + D(v^{l}_o)c^{l}_{t}]		& \text{output gate} \label{output}  \\
	h^{l}_t & = && o^{l}_t \cdot \tanh[c^{l}_t] 
	& \text{cell output} \label{LSTMproj}
	\vspace{-0.2in}
\end{alignat}
The output of the network is given by, $$y_t   = \text{softmax}[W^{L+1}h^{L}_t + b^{L+1}] .$$ 
The peephole connection $D(v^{l}_i)$ is a diagonal matrix with $v^{l}_i$ as its diagonal. We will refer to Equations~\ref{input},~\ref{forget},~\ref{cell},~\ref{output} as input gate, forget gate, cell state and output gate respectively.

 \vspace{-0.15in}
\section{Parameter Reduction Schemes}  \label{sec:related}

We now describe various parameter reduction methods explored in this work, to make the recurrent architectures described in Section \ref{sec:models} more compact. For sake of simplicity, we will refer to the generic matrix we want to compress as $W$, which is of size  $m \times n$. The goal of the compression schemes is to reduce the number of parameters of $W$, namely $mn$.  
 \vspace{-0.1in}
\subsection{Low Rank Factorization}
\label{lowrank} 
In a low-rank compression scheme, we assume that matrix $W$ has rank $r$. Then there exists \cite{strang:matrix} a factorization $W=W_a \times W_b$ where $W_a$ is a full-rank matrix of size $m \times r$ and $W_b$ is a full-rank matrix of size $r \times n$. Thus, we replace matrix $W$ by matrices $W_a$ and $W_b$ which is equivalent to replacing a fully connected layer with a linear bottleneck layer. Notice there is no non-linearity (i.e. sigmoid) between matrices $W_a$ and $W_b$. A low-rank decomposition of Equation~\ref{full} for recurrent and feedforward matrices $U^l$ and $W^l$, for layer $l \in \{1,\ldots,L\}$, is shown in Equation \ref{eq:lowrank}.
\begin{equation}
h^l_t  = \sigma[W_a^{l} W_b^{l}h^{l-1}_t + U_a^l U_b^lh^l_{t-1} + b^l] 
\label{eq:lowrank}
\end{equation}
We can reduce the number of parameters of the system so long as the number of parameters in $W_a$ (i.e., $mr$) and $W_b$ (i.e., $rn$) is less than $W$ (i.e., $mn$). Low-rank matrix factorization was first explored for speech in \cite{tsainath:lowRank}, where it was found DNN models could be reduced by 30-50\% with no loss in accuracy. Furthermore, in the context of DNNs,~\cite{xue2013restructuring} explored computing the singular-value-decomposition (SVD) of a full matrix to learn the two smaller low-rank matrices.
 \vspace{-0.1in}
\subsection{Sharing Low-rank across Layers: Projection Model}\label{sec:projection}

More recently, \cite{Hasim14} shared the low-rank factor across layers of the recurrent architecture. To be more specific, we require $W^{l}_b = U^{l-1}_b$ when we apply low rank to both $U^{l-1}$ and $W^{l}$ as in Equation~\ref{eq:lowrank}.
We can rewrite Equation~\ref{full} for RNNs as
\begin{align}
	h^l_t & = \sigma[W_{a}^{l}m^{l-1}_t + U_{a}^lm^l_{t-1} + b^l]  & 
	 \notag \\
	m^l_t & = U_b^l h^l_t &  l = 1, \ldots, L \label{downproj} \\
	y_t   & = \text{softmax}[W^{L+1}m^{L}_t + b^{L+1}]  & \label{proj}
\end{align}
where $m^0_t= x_t$. $m^l_t$, the low rank output in Equation~\ref{downproj}, can be seen as a linear projection of the original hidden layer, which is shared across layers. We refer to this as a projection compression scheme. 

The projection model is compact with weight sharing. Besides, since the projection weight $U_b^{l-1}$ is shared between $U^{l-1}$ and $W^l$, its gradient also receives error signals from both factors. The error component of $U_b^{l-1}$ coming from $W^l$ is closer to the output, compared to that of $U^{l-1}$, which makes the learning of recurrent connection 
easier in the projection model (Equation~\ref{proj}) compared to the full model (Equation~\ref{full}). Therefore, the projection model regularizes the full model with fewer number of parameters and facilitates learning through weight sharing.

Similarly for LSTM, we can modify Equation~\ref{LSTMproj} to be
\vspace{-0.05in}
\begin{align}
\vspace{-0.05in}
h^{l}_t  =  P^l o^{l}_t \cdot \tanh[c^{l}_t]  &&  \text{projection node}\label{LSTMproj2}
\end{align}
which projects the cell $c_t^l$ down to $h_t^l$ of lower dimensionality, with $P^l$ having a similar interpretation to $U_b^l$ in Equation~\ref{downproj}. Then the projection node $h^{l}_t$ will feed forward to next layer and recurrently to the next time step of the same layer, for all gates and cell activations in Equation~\ref{input}-\ref{output}.  We refer the reader to \cite{Hasim14} for detailed equations regarding the LSTM projection model.
 \vspace{-0.1in}
\subsection{HashedNets}
\label{hash} The HashedNets scheme was recently proposed in~\cite{chen2015compressing} to reduce memory consumption of DNN layers for computer vision tasks.  Here, we assume that the matrix $W$ has only $k$ unique parameter values, instead of $mn$. The connections in $W$ are randomly grouped together and hashed into one bucket of length-$k$ parameter vector $v$, $W_{ij} = v_{h(i,j)}$, where $h(i,j): \mathbb{N}\times\mathbb{N} \rightarrow \{1, \ldots, k\}$ is a predefined hashing function. $v_m$ is shared among all entries of $W_{ij}$ where $h(i,j) = m$ in both feedforward and back-propagation~\cite{chen2015compressing}. 

To be consistent with other methods, we use pseudo rank $r$ for HashedNets to refer to $k=2nr$ number of parameters. We will apply this hashing trick to the $U^l$ and $W^l$ matrices for RNNs. 
 \vspace{-0.1in}
\subsection{Toeplitz-like Structured Matrices}
\label{structure} 
Recently, ~\cite{Sindhwani15} proposed a new family of parameter sharing schemes for small-footprint deep learning based on structured matrices characterized by the notion of {\it displacement operators}~\cite{Pan}.  Unlike HashedNets where weights are randomly grouped, parameter sharing mechanisms in structured matrices are highly specific and deterministic. The structure can be exploited for fast matrix-vector multiplication (forward passes) and also gradient computations during back-propagation typically using Fast Fourier Transform like operations. To get a flavor of this approach, consider Toeplitz matrices where parameters are tied along diagonals.
\vspace{-0.1in}
\begin{eqnarray}
\left[\begin{array}{cccc} 
{\bf \color{red}{t_0}} & {\bf \color{green} t_{-1}} & \ldots & {\bf \color{cyan}{t_{-(n-1)}}}\\
{\bf \color{blue} t_1}& {\bf \color{red} t_0} & \ldots & \vdots\\
\vdots & \vdots & \vdots & {\bf \color{green} t_{-1}} \\
{\bf t_{n-1}} & \ldots & {\bf \color{blue} t_1} & {\bf \color{red}t_0}
 \end{array}
 \right] \nonumber
\end{eqnarray}

It is known that $n\times n$ Toeplitz matrices admit $O(n~\log~n)$ time to compute matrix-vector products, which are essentially equivalent to performing linear convolutions. Toeplitz matrices also have the property that via certain shift and scale operations as implemented by specific displacement operators, they can be linearly transformed into matrices of rank less than or equal to $2$. Thus, the so called {\it displacement rank} of all Toeplitz matrices is up to $2$.  ~\cite{Sindhwani15} propose learning parameter matrices that are generalizations of the Toeplitz structure by allowing the displacement rank $r$ to be higher. This class of matrices are called {\it Toeplitz-like} and they include products and inverses of Toeplitz matrices, and their linear combinations, which can be interpreted as composition of convolutions and deconvolutions. These matrices can be parameterized via a sum over products of $r$ Circulant and Skew-circulant matrices. The displacement rank $r$ serves as a knob on modeling capacity. High displacement rank matrices are increasingly unstructured.~\cite{Sindhwani15} show that on mobile speech recognition problems, such transforms are highly effective for learning compact Toeplitz-like layers compared to fully connected DNNs. 

With displacement rank $r$, there are $2nr$ free parameters in the Toeplitz-like structured matrix. We will apply the Toeplitz-like transform to $U^l, W^l$ in RNNs, and $W_i^l, W^l_f, W^l_c, W^l_o,$ $U_i^l, U^l_f, U^l_c, U^l_o$ to LSTMs in our experiment.

{\bf Summary}: Table~\ref{table:compress} gives a summary of different methods with its number of parameters as a function of an appropriate notion of rank. For simplicity we assume $m=n$. For projection, the number of parameters is averaged across layers where low-rank factors are shared.
\begin{table} [h]
\vspace{-0.1in}
	\centering
	\begin{tabular}{c|c|c}
	\toprule
	method & knob & \# of params \\
	\hline   
	Low-rank & rank $r$ & $2 n r $  \\ 
	Projection & rank $r$ & $\frac{3 }{2}n r$  \\
	HashedNets & pseudo-rank $r$ & $2n r$  \\ 	
	Toeplitz-like &  displacement rank $r$ & $2nr$ \\ 
	\bottomrule
	\end{tabular}
	\caption{comparison of compression methods}
	\label{table:compress}
\vspace{-0.2in}
\end{table}

\vspace{-0.1in}
\section{Experimental Details} \label{sec:exps}
We report two sets of experiments: with RNNs on a medium-sized noisy training set of 300 thousand English-spoken utterances (300 hours), and with LSTMs on a larger training set of 3 million utterances (2,000 hours).  These data sets are created by artificially corrupting clean utterances using a room simulator, adding varying degrees of noise and reverberation such that the overall SNR is between 5dB and 30dB. The noise sources are from YouTube and daily life noisy environmental recordings. 
All training sets are anonymized and hand-transcribed, and are representative of Google's voice search traffic. The training sets is randomly split into 90\% for model training, and 10\% for heldout used to evaluate frame accuracy. WER is reported on a noisy test set containing 30,000 utterances (over 20 hours).

The input feature for all models are 40-dimensional log-mel filterbank features, computed every 10ms. All recurrent layers  are initialized with uniform random weights between  $-0.02$ to $0.02$. The RNNs and LSTMs are unrolled for 20 time steps for training with truncated backpropagation through time (BPTT). In addition, the output state label is delayed by 5 frames, similar to \cite{Sainath15}. 

All neural networks are trained with the cross-entropy criterion, using the asynchronous stochastic gradient descent (ASGD) optimization strategy described in \cite{Dean12}. All networks have 42 phone output targets such that the output layer would have few parameters and we could focus our attention on compressing other layers of the network.  We use a exponentially decaying learning rate, which starts at 0.004 and
has a decay rate of 0.1 over 15 billion frames. We apply gradient clipping from \{1, 10, 100\} in RNN training, and cell clipping at 50 for LSTM training.

 \vspace{-0.1in}
\section{Results} 

\subsection{Learning Compact RNNs}

\subsubsection{Parameter reduction in bottom layers}
We benchmark a full, standard RNN, which has 3 hidden layers with 600 cells per layer, and studied how we can reduce parameters relative to it with various compact architectures. 


We compared different  methods introduced in Section~\ref{sec:related}, with a focus on how bottom layers can be heavily condensed. To be more specific, we compress weight matrices  $U^1, W^2, U^2$ of the first two layers, down to rank $r=5$ for low rank, HashedNets and Toeplitz-like matrices, with the definition of rank shown in Table \ref{table:compress}. Note that $r=5$ is very limited compared to the original 600 dimensionality. 1.85 million parameters of the baseline model is cut down to $\sim$790k, roughly $40\%$ of its original size. 

\begin{table} [h!]
 \vspace{-0.1in}
  \centering
  \begin{adjustbox}{max width=0.5\textwidth}
  \begin{tabular}{c|c|c|c|c}
    \toprule
    model & compression & \# of params & frame accuracy & WER \\
    \hline
    full  & - & 1.85m & 73.26 & 43.5\\ 	\hline   
    low rank 	& $5,5,600$ & 790k &  68.08 & 54.6\\	
    HashedNets  &$5,5,600$ & 790k  & 70.09 & 49.2 \\
    Toeplitz-like  &$5,5,600$ & 790k  & 70.79 & 48.4\\     \bottomrule
  \end{tabular}
  \end{adjustbox}
  \caption{Learning different compact models for RNN}
  \vspace{-0.1in}
  \label{table:compresfull}
\end{table}
From Table~\ref{table:compresfull}, the Toeplitz-like transform is the most efficient to compress bottom layers, which attains the lowest WER under similar number of parameters. Given a fixed budget on model size, different compression schemes make different assumptions while compressing.  The low rank assumption performs the poorest because rank 5 is too constrained. HashedNets imposes a somewhat weaker structure on the parameters via random grouping, and also performs moderately. On the other hand, a Toeplitz-like structured matrix with rank 5 can be interpreted as composition of convolutions and deconvolutions, and performs the best to reduce parameters in the bottom layers.
  \vspace{-0.1in}
 \subsubsection{Parameter reduction across all layers with projection} \label{sec:projection_experiment}
We also benchmarked a low-rank model with shared factors~\cite{Hasim14} as described in~\ref{sec:projection}. Here, we use 
$100$ projection nodes for $m_1, m_2$ and 200 nodes for $m_3$. With just around $635$k parameters,  this model achieves an frame accuracy of $73.72\%$ and word error rate of $43.5\%$ matching the full RNN model with one-third the number of parameters.   This projection model shares weights and thus gradients across all the layers and appears to be a more effective than the  aggressive compression of bottom layers alone using HashedNets, Toeplitz-like matrices or untied low-rank models.    One hypothesis we have is that perhaps Toeplitz-like compression, with its convolution interpretation, is better for lower layers, while shared low-rank factorizations are more effective for higher layers. This experiment was found to be difficult to run with RNNs, 
since in our experiments increasing the number of RNN layers exacerbated the vanishing/exploding gradient problem. Hence, we reserve this set of experiments for the next section, where we compress a deep LSTM with 5 layers, which exhibited more stable optimization behavior for this setting.
 \vspace{-0.1in}
\subsubsection{Displacement rank of Toeplitz-like transfrom}
Next we explore the behavior of Toeplitz-like matrix by changing the displacement rank $r$. In Table \ref{table:st-toeplitz}, the WER improves when we compress $U^1, W^2, U^2$ to higher ranks, at the cost of increasing parameters and training time. From the column of seconds per ASGD optimization step, the training time is proportional to the displacement rank of Toeplitz-like matrix. As rank 5 gives us a reasonable tradeoff between performance and training time, we will use it for further structured matrix experiments.


\begin{table} [h!]
	\centering
	\vspace{-0.1in}
	\begin{tabular}{c|c|c|c|c}
	\toprule
	rank & \# of params & sec. per step & frame accuracy & WER\\
	\hline   
	2 &  779k &  0.13 & 70.41 & 49.6 \\     
	5 &  790k & 0.28 & 70.79 & 48.4 \\ 	
	10& 808k & 0.50 & 70.99 & 47.9 \\ 
	\bottomrule
	\end{tabular}
	\caption{Toeplitz-like matrices with different rank }
	\label{table:st-toeplitz}
	\vspace{-0.2in}
\end{table}

 \vspace{-0.1in}
\subsection{Learning Compact LSTMs}
In light of the observations from the RNN experiments, we mainly focus on LSTM with projection layer (Equation~\ref{LSTMproj2}) and Toeplitz-like transforms in bottom layers for compact LSTM experiments.
The full LSTM model has 5 hidden layer, with 500 hidden unit per layer. For projection model, we introduce projection $P^1-P^5$, as in Equation~\ref{LSTMproj2}, to all hidden layers, and the two numbers in column ``compression'' of Table~\ref{table:lstm} indicate the dimensions of $h^1$ to $h^4$, and $h^5$ respectively. 
For Toeplitz-like transform, we start from a projection model with 100 nodes in $h^1$ to $h^4$, and 200 nodes in $h^5$, and replace weight matrices with Toeplitz-like matrices progressively. `$U_\text{x}^j$' in column ``compression''  stands for that all four weights $U_i^j, U_f^j, U_c^j, U_o^j$ in recurrent connections of layer $j$ are compressed. We reduce all  gates and cell state equally. Note that when we compress $U_\text{x}^{l-1}$ and $W_\text{x}^l$ at the same time, projection layer $P^{l-1}$ will be removed. Column ``projection'' details the projection layers in the model.

\begin{table} [h!]
  \centering
  \vspace{-0.1in}
  \begin{tabular}{c|c|c|c|c}
    \toprule
    model & proj. &compression &  \# of params & WER \\
    & & $h^1 -h^4$, $h^5$ & & \\ 
    \hline
    full 	& - 		&  - & 9.12m          		& 33.1   \\  \hline
    proj.   	& $P^1-P^5$ & 100, 200 &  2.41m  		&  33.6 \\  
    proj. 	 & $P^1-P^5$ &   90, 200  &  2.23m    & 33.8    \\
    proj.  & $P^1-P^5$   & 80, 200      & 2.05m  &  	34.2 \\
    \hline
    Toep.   & $P^1-P^5$ & $U_\text{x}^1$ & 2.23m  & 33.4 \\
    Toep.   & $P^2-P^5$ & $U_\text{x}^1, W_\text{x}^2$ & 2.00m  & 33.5 \\
    Toep.    & $P^2-P^5$ & $U_\text{x}^1, W_\text{x}^2, U_\text{x}^2$ & 1.82m   & 33.9 \\
    Toep.   & $P^3-P^5$ & $U_\text{x}^1, W_\text{x}^2, U_\text{x}^2, W_\text{x}^3$ & 1.59m  & 35.4 \\		
    \bottomrule
  \end{tabular}
  \caption{Learning different compact models for LSTM}
  \vspace{-0.15in}
  \label{table:lstm}
\end{table}

For both projection model and Toeplitz-like transform, the performance drops as we reduce the number of parameters. However, we see that given a fixed model size, it is more effective to compress lower layers with Toeplitz-like matrices of low displacement rank, which has more of a convolutional interpretation, compared to projection compression on all layers with moderate rank. We see this is quite different than the behavior of compression on a shallow RNN (Section ~\ref{sec:projection_experiment}), since we can afford to have a deeper network with LSTM . Overall, with the combination of Toeplitz-like transform and projection, we are able to compress the LSTM model down by 75\% to 2.2m parameters, with only a 0.3\% increase in WER to 33.4.

Next, we try to answer the question of where in the LSTM we should apply the compression. 
We compare the effect of compressing recurrent weight $U^l$ or non-recurrent weight $W^l$ in Table~\ref{table:lstmproj}. ``compression'' lists all Toeplitz-like matrices, where all 4 gates are compressed equally, and ``projection'' specifies projection layers in the model.  
As we can see, it makes no significant difference whether we compress recurrent or non-recurrent weight as long as the number of parameter matches.

\begin{table} [h!]
  \centering
  \vspace{-0.1in}
  \begin{tabular}{c|c|c|c}
    \toprule
    compression & projection & \# of params & WER \\
    \hline
    $U_\text{x}^1, W_\text{x}^2$  & $P^1 - P^5$ & 2.05m &  33.8 \\
    $U_\text{x}^1, U_\text{x}^2$  & $P^1 - P^5$ & 2.05m  &  33.5 \\
    \hline
    $U_\text{x}^1, W_\text{x}^2, U_\text{x}^2, W_\text{x}^3$  &  $P^2 - P^5$ &1.64m  &   34.8 \\	
    $U_\text{x}^1, W_\text{x}^2, U_\text{x}^2, U_\text{x}^3$  &  $P^2 - P^5$ & 1.64m  &  34.7 \\			
    \bottomrule
  \end{tabular}
  \caption{Compression of feedforward or recurrent weights in LSTM}
	 \vspace{-0.15in}
  \label{table:lstmproj}
\end{table}

Thus far we have compressed all gates in the same way. However, different gates have different levels of importance in LSTM~\cite{greff2015lstm} and perhaps less important gates can be compressed more. Thus, we investigate how the WER is affected by reducing different gates differently with Toeplitz-like matrices. 

We take a LSTM projection model, which has 100 projection nodes in $h^2$ to $h^4$, and 200 projection nodes in $h^5$. All gates and cell states of $U^1, W^2, U^2$ are Toeplitz-like matrices. We vary the compression on cell state and gates of $W^3$ with Toeplitz-like transforms, and record the change in WER. These models lie in between the models of last two rows in Table~\ref{table:lstm}, which has a huge jump in WER from 33.9 to 35.4.

In Table~\ref{table:lstmgates}, gates column indicates if input gate, forget gate, cell state or output gate of $W^3$, 
is being compressed. 
Compressing the cell state (Equation~\ref{cell}) makes a major difference, 0.4\% in WER, for LSTM performance. Compressing forget gate or not alone does not show much impact with 0.1\% increase in WER.  
But reducing input gate, output gate and forget gate altogether, with 0.14m fewer parameters, would make 0.5\% worse in WER.
Both~\cite{greff2015lstm,zaremba2015empirical} notice that removing forget gate would significantly hurt the performance, and ~\cite{zaremba2015empirical} finds that output gate is of the least importance. We do not identify significant difference in compressing different gates in our experiment, probably because we only change the gates for the third layer input weights in a five-layer LSTM, and that rank 5 Toeplitz-like matrices are sufficient to retain enough information for most gates.

\begin{table} [h!]
\vspace{-0.15in}
	\centering
	\begin{adjustbox}{max width=0.5\textwidth}
	\begin{tabular}{c|c|c|c|c}
		\toprule
		proj.  & compression & compressed gates & \# of params  & WER \\
		\hline		
		 $P^2 - P^5$ & $U_\text{x}^1, W_\text{x}^2, U_\text{x}^2$  & $W^3_i, W^3_f, W^3_c, W^3_o$ & 1.64m  &   34.8 \\	
		 $P^2 - P^5$ & $U_\text{x}^1, W_\text{x}^2, U_\text{x}^2$  & $W^3_i, W^3_f, W^3_o$ &1.68m  & 34.4 \\
		 $P^2 - P^5$ & $U_\text{x}^1, W_\text{x}^2, U_\text{x}^2$  & $W^3_i, W^3_o$ &  1.73m  & 34.3  \\
		 $P^2 - P^5$ & $U_\text{x}^1, W_\text{x}^2, U_\text{x}^2$  & $-$  & 1.82m   & 33.9 \\
		\bottomrule
	\end{tabular}
	\end{adjustbox}
	\caption{Compression of different gates in LSTM}
	\vspace{-0.2in}
	  \label{table:lstmgates}
\end{table}

 \vspace{-0.1in}
\section{Conclusions}
In this work, we studied how to build compact recurrent neural networks for LVCSR tasks.  In our RNN experiments, we noted that Toeplitz-like structured matrices outperform HashedNets and Low-rank bottleneck layers for aggressive parameter reduction in the bottom layers.  For LSTM parameter reduction,  architecting upper layers with projection nodes to moderate rank, and bottom layers with Toeplitz-like transforms was found to be a particularly effective strategy. With this strategy, we are able to build a compact model with 75\% fewer parameters than a standard LSTM model, while only incurring 0.3\% increase in WER. Compressing recurrent or non-recurrent weight does not make significant difference.  We find that LSTM performance is sensitive to cell state compression, making a noticeable change in WER. 

\section{Acknowledgements} 
Thank you to Rohit Prabhavalkar, Ouais Alsharif and Hasim Sak for useful discussions related to model compression and LSTMs. 
\bibliographystyle{IEEEbib}
\bibliography{main}
\end{document}